\documentclass[letterpaper]{article} 
\usepackage{aaai2026}  
\usepackage{times}  
\usepackage{helvet}  
\usepackage{courier}  
\usepackage[hyphens]{url}  
\usepackage{graphicx} 
\usepackage{natbib}  
\usepackage{caption} 
\usepackage{algorithm}
\usepackage{algorithmic}

\usepackage{newfloat}
\usepackage{listings}
\DeclareCaptionStyle{ruled}{labelfont=normalfont,labelsep=colon,strut=off} 
\floatstyle{ruled}
\newfloat{listing}{tb}{lst}{}
\floatname{listing}{Listing}
\usepackage{booktabs}
\usepackage{colortbl}
\usepackage{url}
\usepackage{blindtext}
\usepackage{multirow}
\usepackage[utf8]{inputenc} 
\usepackage[T1]{fontenc} 
\usepackage{amsfonts} 
\usepackage{nicefrac} 
\usepackage{microtype} 
\usepackage{xcolor} 
\usepackage{bbding}
\usepackage{soul}
\usepackage{pifont}
\usepackage{amsmath}
\usepackage{enumitem}
\usepackage{xspace}

\newcommand{\red}[1]{\textcolor[rgb]{1,0,0}{#1}}

\newcommand{\green}[1]{\textcolor[rgb]{0,0.69,0.31}{#1}}
\definecolor{officegreen}{rgb}{0.0, 0.5, 0.0}

\newcommand{\im}{\mathbf{I}}
\newcommand{\imout}{\hat{\mathbf{I}}}
 
\newcommand{\gaze}{\mathbf{g}}

\newcommand{\hen}{\mathcal{F}_{h}}
\newcommand{\len}{\mathcal{F}_{l}}

\newcommand{\lfe}{z_{l}}
\newcommand{\hfe}{z_{h}}

\def\ie{\emph{i.e.}\xspace}

\title{RTGaze: Real-Time 3D-Aware Gaze Redirection from a Single Image}
\author{
    Hengfei Wang\textsuperscript{\rm 1},
    Zhongqun Zhang\textsuperscript{\rm 1,2},
    Yihua Cheng\textsuperscript{\rm 1,}\corresponding,
    Hyung Jin Chang\textsuperscript{\rm 1}
}
\affiliations{
    \textsuperscript{\rm 1}University of Birmingham\\
    \textsuperscript{\rm 2}College of Software, Nankai University\\

    hengfei\_wang@163.com, zhangzhongqun@nankai.edu.cn, \{y.cheng.2, h.j.chang\}@bham.ac.uk
}

\usepackage{bibentry}

\begin{document}

\maketitle

\begin{abstract}
    Gaze redirection methods aim to generate realistic human face images with controllable eye movement. However, recent methods often struggle with 3D consistency, efficiency, or quality, limiting their practical applications.
    In this work, we propose RTGaze, a real-time and high-quality gaze redirection method. Our approach learns a gaze-controllable facial representation from face images and gaze prompts, then decodes this representation via neural rendering for gaze redirection. Additionally, we distill face geometric priors from a pretrained 3D portrait generator to enhance generation quality.
    We evaluate RTGaze both qualitatively and quantitatively, demonstrating state-of-the-art performance in efficiency, redirection accuracy, and image quality across multiple datasets. Our system achieves real-time, 3D-aware gaze redirection with a feedforward network (~0.06 sec/image), making it 800× faster than the previous state-of-the-art 3D-aware methods.
\end{abstract}

\section{Introduction}
\label{sec:intro}

Gaze is one of the most important facial features and it conveys human attention and intention in interaction. 
Gaze redirection involves redirecting the gaze of a face image to a given target direction without changing the identity.
It has various applications including virtual reality \cite{pai2016gazesim, mania2021gaze, zhang2022trans6d, zheng2023hs}, digital human \cite{jack2015human, tse2022s, choi2025roll, wang2023high} and CG film-making \cite{yang2022recursive, blanz2004exchanging}. 

Existing gaze redirection methods can be broadly divided into two categories: 2D-based and 3D-based, depending on whether they incorporate 3D representations.
2D-based methods achieve gaze redirection either by warping pixels in the input image \cite{Ganin_2016_ECCV} or by generating new gaze images through deep generative models such as Generative Adversarial Networks (GANs) \cite{He_2019_ICCV, jindal2023cuda}, encoder-decoder networks \cite{Park_2019_ICCV}, and Variational Autoencoders (VAEs) \cite{zheng2020self}. While effective to some extent, these methods do not capture the inherently 3D nature of gaze redirection, resulting in suboptimal performance under larger head poses.

\begin{figure}[t]
    \centering
    \includegraphics[width=1.0\linewidth]{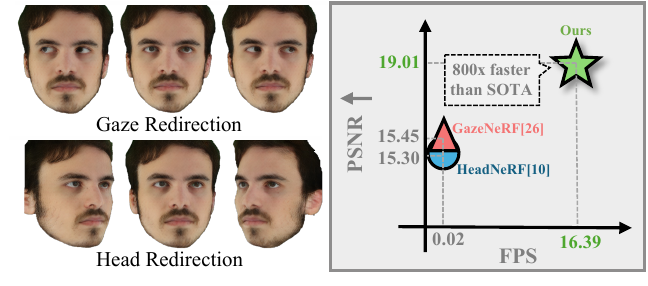}
    \caption{
    \textbf{3D-aware gaze redirection results} from our proposed RTGaze, which generates photo-realistic face images under novel gazes and views with good 3D consistency in real time. Compared to the state-of-the-art 3D-aware gaze redirection method GazeNeRF \cite{ruzzi2022gazenerf}, which requires approximately one minute during inference, our approach achieves real-time performance at $\textbf{61ms}$ while maintaining superior image quality.
    }
    \label{fig:teaser}
\end{figure}

3D-based methods, on the other hand, construct a 3D representation of each input face image using techniques like the neural radiance field (NeRF) \cite{mildenhall_2020_nerf}. Once trained, these models can generate a full 3D face and, by adjusting camera poses, produce images with varied head orientations, ensuring strong 3D consistency across a wide range of poses. Among these, GazeNeRF \cite{ruzzi2022gazenerf} is the state-of-the-art, employing two separate multilayer perceptrons (MLPs) to model the radiance fields for the face and eyes independently. GazeNeRF generates novel views using latent codes and gaze labels, but during inference, it requires GAN inversion and updating learnable latent codes before rendering \cite{hong2022headnerf, ruzzi2022gazenerf}, a process that is time-consuming and delays gaze redirection.
Balancing 3D consistency with real-time performance, therefore, remains an open challenge in gaze redirection.

\begin{figure*}[t]
    \centering
    \includegraphics[width=0.9\linewidth]{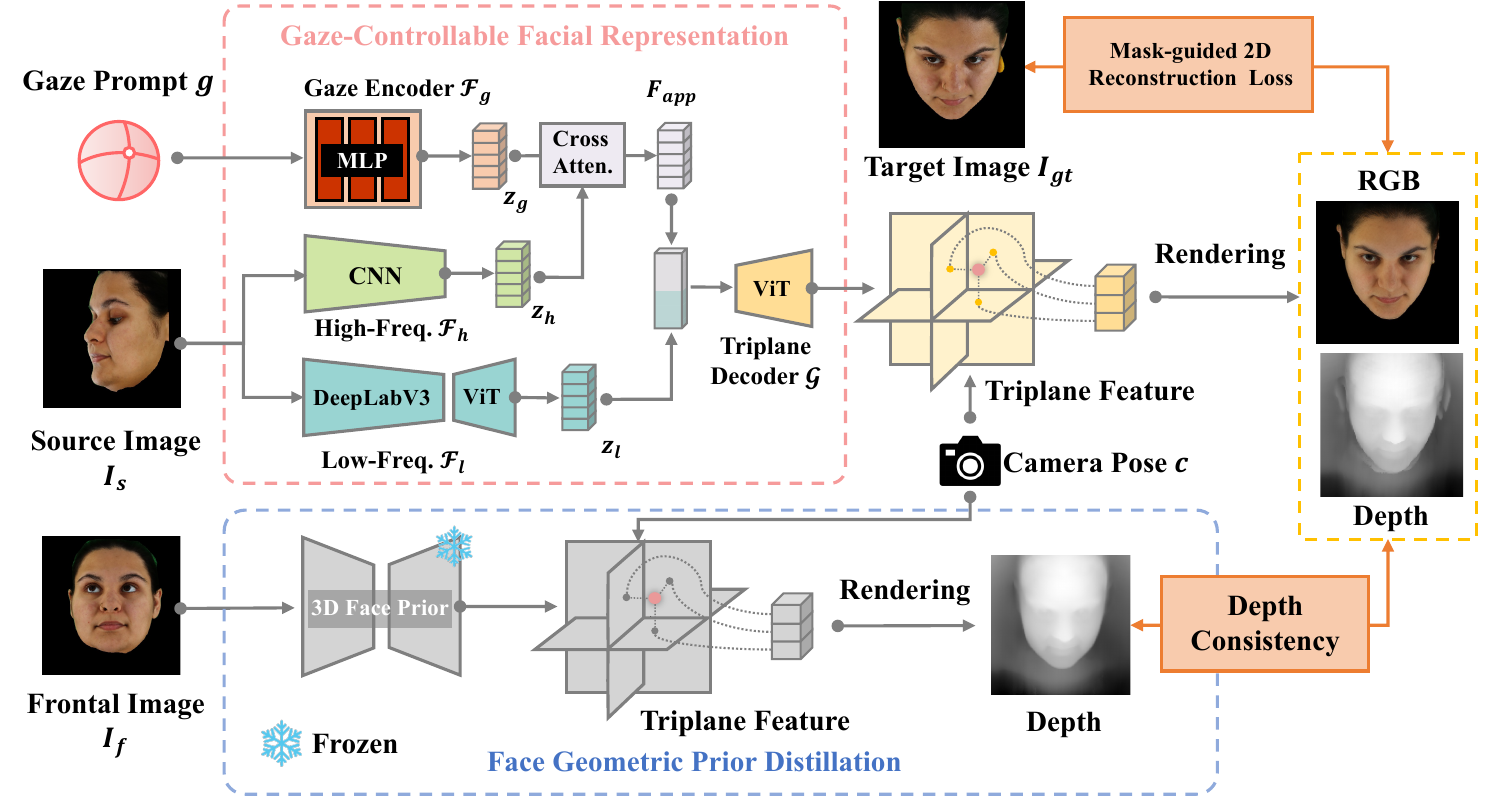}
    \caption{
    Our model takes three inputs: gaze prompts, source images, and frontal images during training. It consists of a gaze-controllable facial representation learning module and face geometric prior distillation module. First, the model extracts high-frequency and low-frequency features from the source images and injects the gaze prompt into the high-frequency features. The final representation is a fusion of the injected gaze features and the low-frequency features. This combined representation is then fed into a triplane decoder, which generates a 3D face representation in the form of a triplane. This triplane representation is used to render the final gaze-redirected image. The target image provides a mask-guided 2D constraint along the eye region. Additionally, we aim to distill 3D face geometry prior from a pre-trained 3D portrait generation model. We compute depth images from both the pre-trained model and our model, and apply a distillation loss.
     }
    \label{fig:pipeline}
\end{figure*}

In this paper, we introduce RTGaze, a novel method for real-time gaze redirection with 3D awareness, achieving both high-efficiency and high-quality generation. 
As illustrated in Fig. \ref{fig:pipeline}, our method takes images and gaze prompts as inputs, learns a gaze-controllable facial representation. We employ two distinct encoders to separately extract high-frequency and low-frequency features from images. The gaze prompt is injected via a cross-attention mechanism, merging it directly with the high-frequency features, which are later fused with the low-frequency features. 
On the other hand, directly optimizing appearance and shape in a lightweight model is challenging, particularly when inferring 3D geometry from a single image \cite{liu2023zero, liu2024one}. To address this, we distill prior knowledge of facial geometry from a pre-trained 3D portrait generator into our module. Our method utilizes NeRF's 3D structure learning and applies a distillation loss to the learned geometric depth images.

Our system achieves \textit{real-time} 3D-aware gaze redirection through a feedforward network, processing each frame in just $\mathbf{61ms}$ on a standard consumer GPU. Extensive quantitative and qualitative evaluations validate our approach, and a series of ablation studies confirm the effectiveness of our design choices. Compared with existing methods \cite{ruzzi2022gazenerf, hong2022headnerf}, RTGaze offers superior image quality and redirection accuracy with a significant boost in inference speed. 
In summary, our contributions are as follows:

\begin{enumerate}[itemsep=2pt,topsep=0pt,parsep=0pt]

\item We present a real-time 3D-aware gaze redirection model that achieves both high-efficiency and high-quality gaze-controllable image generation. Our method surpasses state-of-the-art methods in inference speed, redirection accuracy, and image quality across different datasets.

\item We propose a novel module for learning gaze-controllable facial representation from face images and a gaze prompt. It consists of two distinctive encoders for extracting facial features and a gaze injection module to effectively incorporate the gaze prompt into the facial representation.

\item We introduce the distillation of 3D face priors from a 3D portrait generation network into our gaze redirection model. By applying a distillation loss on the learned geometric depth images, our approach enhances the overall quality of gaze-redirection synthesis.

\end{enumerate}

\section{Related Work}
\label{sec:related}
Gaze redirection methods can generally be categorized into 2D-based methods and 3D-based methods, depending on whether they incorporate 3D representations.

\subsection{2D-based Gaze Redirection}
Deepwarp \cite{Ganin_2016_ECCV} employs warping maps learned from pairs of eye images with different gaze directions, which requires extensive annotated data.  To reduce this reliance on annotated real data, \citet{Yu_2019_CVPR} incorporate a pretrained gaze estimator with synthetic eye images, further refined by \citet{Yu_2020_CVPR} with an unsupervised gaze representation learning network. 
GAN-based approaches~\cite{He_2019_ICCV} enable gaze redirection by leveraging generative models. FAZE introduces an encoder-decoder framework that encodes eye images into latent vectors, which are then manipulated with rotation matrices to produce synthetic images featuring redirected gaze. ST-ED \cite{zheng2020self} builds on this by disentangling latent representations to perform both head and gaze redirection for full-face images, achieving highly accurate results. ReDirTrans \cite{jin2023redirtrans} projects edited embeddings back into the original latent space, allowing for attribute replacement with minimal impact on other features and preserving the latent distribution. 2D-based methods often struggle with 3D consistency, as they lack an explicit 3D facial representation.

\subsection{3D-based Gaze Redirection}
3D-based gaze redirection methods offer improved 3D consistency via learned 3D representation. EyeNeRF \cite{li2022eyenerf} combines explicit surface modeling for the eyeball with implicit volumetric representations of surrounding eye structures, enabling high-fidelity gaze redirection with photorealistic effects using a minimal setup of lights and cameras. GazeNeRF \cite{ruzzi2022gazenerf} employs a two-stream MLP architecture to separately model the face and eye regions via neural radiance fields, allowing for independent manipulation of the eyeball orientation. HeadNeRF \cite{hong2022headnerf} can be adapted for gaze redirection by integrating gaze labels as conditional inputs. Despite their robust 3D consistency, these methods often require complex, resource-intensive models, limiting their real-time applicability. 

\section{Method}

\subsection{Preliminary}

This work introduces RTGaze, a novel method for real-time gaze redirection from a single image. Given an input image 
$\im$ and a target gaze direction $\gaze$, RTGaze generates a new face image $\imout$ with the specified gaze. 
Our pipeline consists of three main components. First, we build a feature extractor to obtain a gaze-controllable facial representation $f$ from $\im$ and $\gaze$.
Then, we feed $f$ into a decoder $\mathcal{G}$ to generate the triplane representation $\mathbf{T}$~\cite{chan2022efficient}. Finally, we perform neural rendering $\mathcal{N}$ based on $\mathbf{T}$ to synthesize images under the target camera pose. The process could be formulated as:
\begin{equation}
    f=\mathcal{F}\left(\im, \gaze\right), \quad \mathbf{T} = \mathcal{G}\left(f\right), \quad \hat{I} = \mathcal{N}\left(\mathbf{T}, \mathbf{c}\right)
\end{equation}

RTGaze employs a hybrid facial representation, leveraging two distinct networks to separately learn high-frequency and low-frequency facial features. To effectively incorporate gaze control, we design a gaze injection module that integrates the gaze input into the hybrid facial representation.  Additionally, we distill 3D geometry priors from a pre-trained 3D portrait generation model to enhance the quality of gaze redirection.

\subsection{Gaze-Controllable Facial Representation}

Given an input image $\im$, we construct a hybrid image encoder comprising a high-frequency feature encoder $\hen$ and a low-frequency feature encoder $\len$ \cite{trevithick2023real}. The high-frequency encoder captures fine-grained appearance details, while the low-frequency encoder extracts global geometric information.
We feed $\im$ into $\hen$ and $\len$, producing the corresponding high-frequency feature $\hfe$ and low-frequency feature $\lfe$, respectively.
In detail, We employ the DeepLabV3 network~\cite{chen2019rethinking}, pre-trained on ImageNet~\cite{deng2009imagenet}, to capture global contextual and semantic information. This information is then processed by a vision transformer encoder~\cite{trevithick2023real}, which leverages self-attention mechanisms to refine the extracted global features, producing the final low-frequency feature representation.
Simultaneously, a convolutional neural network (CNN) is used to extract high-frequency features, focusing on fine details within the input image.

\noindent\textbf{Gaze Prompt Injection.}
Our aim is to inject the gaze prompt into the hybrid facial representation to obtain a gaze-controllable facial representation. The gaze serves as a prompt for generating the target image, ensuring accurate eye appearance. 
The gaze prompt $\mathbf{g}\in\mathbb{R}^2$ consists of pitch and yaw angles of eyeball rotation.
Since gaze redirection primarily affects eye appearance rather than geometry, our strategy first injects the gaze prompt into the high-frequency feature. The injected feature is then fused with the low-frequency feature to produce the final gaze-controllable facial representation.
More concretely, given the gaze prompt $\gaze$, we first embed it using an MLP layer, ensuring that the embedding length matches that of the high-frequency feature.
Next, we employ a cross-attention layer~\cite{Rombach_2022_CVPR} to inject the gaze embedding into the high-frequency feature. In this process, the high-frequency feature serves as the query, while the gaze embedding acts as both the key and value.
Finally, we integrate the injected high-frequency feature with the low-frequency feature to obtain the final facial representation.

\subsection{Face Geometric Prior Distillation}

Directly optimizing appearance and shape from a single image is challenging, as it lacks sufficient constraints to accurately represent a 3D scene. In this work, we distill face geometric priors from a pre-trained model to enhance the quality of the generated images. Notably, the pre-trained model is typically not designed for gaze redirection. 

In detail, we distill face geometric prior from a pre-trained 3D portrait generation model~\cite{trevithick2023real}. We find the model has the best performance when the face in the input image is oriented frontally.
Therefore, we input a frontal image whose identity and gaze are same as the target image to the pre-trained model and obtain a NeRF representation, \ie, the triplane feature $\mathbf{T}$.
The triplane feature enables the rendering of multi-view images and dense depth maps by predicting the color and density of 3D points along camera rays.
In this work, we focus solely on depth map generation and apply a distillation loss on the predicted depth, as the synthesized images do not always maintain appearance consistency with the target images.
Specifically, we sample 3D points along the camera rays $\mathbf{r}$ under the target camera pose $\mathbf{c}$. For each ray $\mathbf{r}$, the depth value $D(\mathbf{r})$ is computed as:

\begin{equation}
        D(\mathbf{r}) = \sum_{i=1}^N W_i\left(1-\exp \left(-\sigma_i \delta_i\right)\right) \mathbf{d}_i,
 \end{equation}
and
\begin{equation}
        W_i = \exp \left(-\sum_{j=1}^{i-1} \sigma_j \delta_j\right).
\end{equation}

\noindent $N$ is the number of samples along the ray, $\mathbf{d}_i$ is the distance between the $i$-th sample and the camera, $\sigma_i$ is the density of the $i$-th sample, and $\delta_i$ is the distance between the $i$-th and $(i+1)$-th samples.
The depth map $\mathbf{D}$ under the camera pose $\mathbf{c}$ is obtained by integrating all depth values.
We compute the teacher depth map $\mathbf{D^{t}}$ from the pre-trained model and the student depth map $\mathbf{D^{s}}$ from our model, applying an L1 loss to enforce depth consistency between them.
  
\begin{equation}
\mathcal{L}_{\mathcal{D}}=\left\|\mathbf{D}^t-\mathbf{D}^s\right\|_1.
\end{equation}

\subsection{Training Objectives}
We train the model using a pair of images $\mathbf{I}$ and $\mathbf{I}_{t}$ with the same identity but different gazes.
The 3D face prior is obtained from an additional frontal image with the same identity and gaze as $\mathbf{I}_{t}$.
We optimize our model using the following objective function:
\begin{equation}
\mathcal{L}=\alpha \cdot \mathcal{L}_{\mathcal{R}}+\beta \cdot \mathcal{L}_{\mathcal{D}}+\gamma\cdot\mathcal{L}_{\mathcal{P}},
\label{eq:total_loss}
\end{equation}
where $\mathcal{L}_{\mathcal{R}}$, $\mathcal{L}_{\mathcal{D}}$, $\mathcal{L}_{\mathcal{P}}$ represent the reconstruction loss, distillation loss, and perceptual loss, respectively.

\noindent \textbf{Mask-Guided 2D Reconstruction Loss.}
We apply a reconstruction loss that minimizes the differences between the generated image $\hat{\mathbf{I}}$ and the target image  $\mathbf{I}_{t}$ in pixel level.
To improve the eye generation quality, we apply an eye region mask to the reconstruction loss and enhance eye region reconstruction by applying a greater loss coefficient specifically to the eye area.
\begin{equation}
\mathcal{L}_{\mathcal{R}}=\alpha_1 \cdot \mathcal{L}^{face}_{\mathcal{R}} + \alpha_2 \cdot \mathcal{L}^{eye}_{\mathcal{R}},
\label{eq:recon_loss}
\end{equation}
where $\mathcal{L}^{face}_{\mathcal{R}}$ and $\mathcal{L}^{eye}_{\mathcal{R}}$ stand for face region reconstruction loss  (excluding the eyes)  and eye region reconstruction loss respectively.
The $\mathcal{L}^{face}_{\mathcal{R}}$ is formulated as:
\begin{equation}
\mathcal{L}^{face}_{\mathcal{R}}=\frac{1}{\left|\mathbf{M}_f \odot \mathbf{I}_{t}\right|}\left\|\mathbf{M}_f \odot\left(\hat{\mathbf{I}}-\mathbf{I}_{t}\right)\right\|_1,
\end{equation}
where $\mathbf{M}_{f}$ is the face region mask and $\odot$ denotes the pixel-wise Hadamard product operator.
$\mathcal{L}_{\mathcal{R}_e}$ is formulated as:
\begin{equation}
\mathcal{L}^{eye}_{\mathcal{R}}=\frac{1}{\left|\mathbf{M}_{e} \odot \mathbf{I}_{t}\right|}\left\|\mathbf{M}_{e} \odot\left(\hat{\mathbf{I}}-\mathbf{I}_{t}\right)\right\|_1,
\end{equation}
where $\mathbf{M}_{e}$ is the eye region mask, and $\mathbf{M}_f = \mathbf{1} - \mathbf{M}_e$.

\noindent \textbf{Perceptual Loss.}
We utilize a perceptual loss \cite{johnson2016perceptual} function to ensure perceptual alignment between the gaze-redirected image $\hat{\mathbf{I}}$ and the target image $\mathbf{I}_{t}$:
\begin{equation}
\mathcal{L}_{\mathcal{P}}=\sum_i \frac{1}{\left|\phi_i\left(\mathbf{I}_{t}\right)\right|}\left\|\phi_i(\hat{\mathbf{I}})-\phi_i\left(\mathbf{I}_{t}\right)\right\|_1,
\end{equation}
where $\phi_i$ denotes the $i$-th layer of a VGG16 \cite{simonyan2014very} network pre-trained on ImageNet \cite{krizhevsky2012imagenet}. 

During inference, our model only takes a single 2D portrait image and a specified gaze direction as input and produces a gaze-redirected, triplane-based 3D face NeRF. 
Our model enables photorealistic view synthesis, allowing for highly realistic visualizations from multiple perspectives.

\begin{table*}[t]
    \centering
    \setlength{\tabcolsep}{5.5pt}
    \caption{Quantitative comparisons on ETH-XGaze dataset. We compare our model with state-of-the-art methods regarding image quality (SSIM, PSNR, LPIPS, FID) and inference speed
    (Encode Time, Render Time, Total Time). For fairness, we only report inference
    speed metrics for 3D methods. Image quality is evaluated on the personalized
    test set from ETH-XGaze, while inference speed is averaged over 100 samples
    on a single NVIDIA 3090 GPU. RTGaze achieves real-time performance,
    processing each image in 61ms, outperforming other methods in most quality
    metrics and maintaining competitive SSIM scores. 
    }
    \begin{tabular}{ccccccccc}
        \hline
        \textbf{Method} & \textbf{3D-based} & \textbf{FID} $\downarrow$ & \textbf{PSNR} $\uparrow$ & \textbf{LPIPS} $\downarrow$ & \textbf{SSIM} $\uparrow$ & \textbf{Enc Time} $\downarrow$ & \textbf{Rend Time} $\downarrow$ & \textbf{Total Time} $\downarrow$ \\
        \hline
        ST-ED & \red{\XSolidBrush} & $115.020$ & $17.530$ & $0.300$ & $0.726$ & - & - & - \\
        HeadNeRF & \green\Checkmark & $69.487$ & $15.298$ & $0.294$ & $0.720$ & $60s$ & $0.058s$ & $60.058s$ \\
        GazeNeRF & \green\Checkmark & $81.816$ & $15.453$ & $0.291$ & $\textbf{0.733}$ & $60s$ & $0.060s$ & $60.060s$ \\
        \rowcolor[rgb]{0.902,0.902,0.902} RTGaze (ours) & \green\Checkmark & $\textbf{38.346}$ & $\textbf{19.007}$ & $\textbf{0.262}$ & $0.715$ & $\textbf{0.026s}$ & $\textbf{0.035s}$ & \textbf{0.061s} \\
        \hline
    \end{tabular}
    \label{tab:sota}
\end{table*}

\begin{figure*}
    \centering
    \includegraphics[width=1\linewidth]{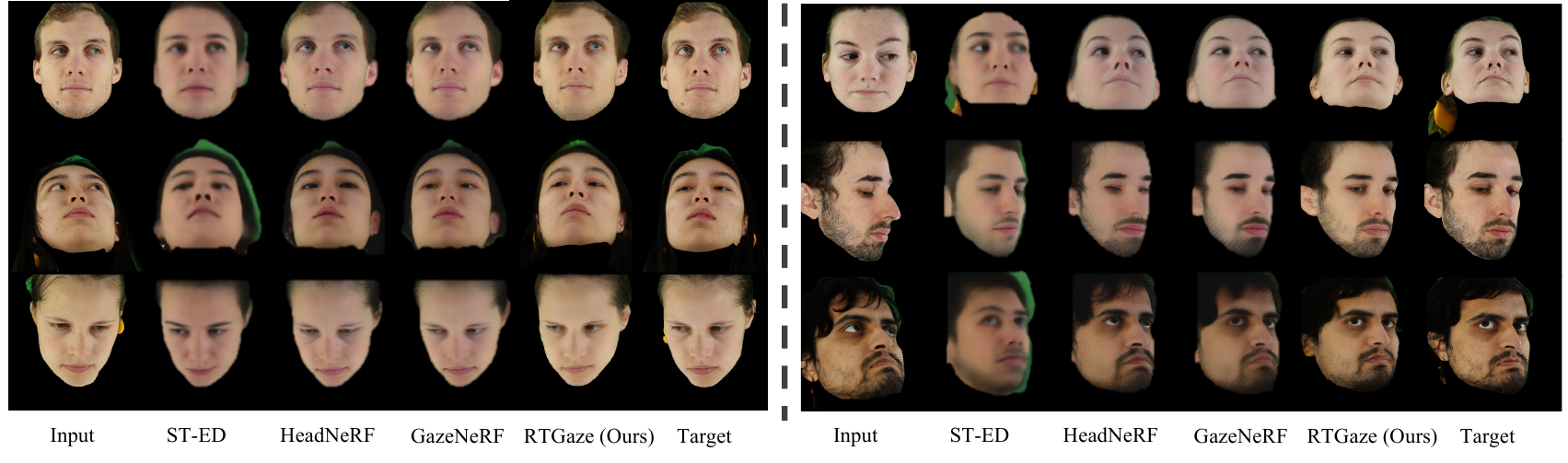}
    \caption{Qualitative comparisons on ETH-XGaze dataset. The background is removed by applying face masks. The
    images generated from RTGaze are photo-realistic and have extensive details.
    ST-ED \cite{zheng2020self} struggles to preserve identity
    information while retaining the unmasked green background which is not found
    in 3D-based methods. HeadNeRF \cite{hong2022headnerf} and GazeNeRF \cite{ruzzi2022gazenerf}
    suffer from losing facial details. }
    \label{fig:sota}
\end{figure*}

\section{Experiments}

\subsection{Datasets} 

\textbf{ETH-XGaze} \cite{Zhang_2020_ECCV} is a large-scale gaze dataset with high-resolution images covering diverse head poses and gaze directions. Collected via a multi-camera setup under various lighting conditions, it includes 756K frames from 80 subjects, each frame captured from 18 angles. A personalized test set comprises 15 subjects, each contributing 200 images with accurate gaze labels. Following GazeNeRF \cite{ruzzi2022gazenerf}, we train RTGaze on 14.4K images from 10 frames per subject, with 18 views per frame, using the ETH-XGaze training set. Models are tested on the personalized test set.

\noindent \textbf{MPIIFaceGaze} \cite{Zhang_2017_CVPRW} is an extension of the MPIIGaze dataset \cite{Zhang_2015_CVPR}, designed for appearance-based gaze estimation. This dataset comprises 3000 facial images, each annotated with two-dimensional gaze labels for a total of 15 subjects.

\noindent \textbf{ColumbiaGaze} \cite{Smith_2013_UIST} comprises 5880 high-resolution images collected from 56 subjects. For each subject, the images were captured using five consistent head poses, each linked to 21 fixed gaze directions.

\subsection{Experimental Setup}

\textbf{Data Preparation.} We first normalize the data and resize the face images
into a resolution of 512x512 following the method provided in ETH-XGaze \cite{Zhang_2020_ECCV}.
Then we process the normalized data following EG3D \cite{chan2022efficient} to
get the camera pose for each image. To realize the mask-guided 2D constraint, we
generate face region masks and eye region masks with face parsing models
\cite{yu2018bisenet}. We convert the provided gaze labels into pitch-yaw labels in
the head coordinate system for convenience of gaze controlling in 3D space.

\noindent \textbf{Implementation Details.} Our model is trained in an end-to-end manner. We
employ Adamw \cite{loshchilov2017decoupled} as our optimizer for whole model The
learning rates are set to $1 e^{-5}$ and $1 e^{-5}$ for the encoding part and the
rendering part respectively. We train our model with a batch size of 4 for 50
epochs. We empirically set the loss coefficients ($\mathcal{L}_{\mathcal{R}}$, $\mathcal{L}
_{\mathcal{D}}$, $\mathcal{L}_{\mathcal{P}}$) in equation \eqref{eq:total_loss}
to $1$, $1$, $0.8$ respectively. The coefficients ($\mathcal{L}^{face}_{\mathcal{R}}$,  $\mathcal{L}^{eye}_{\mathcal{R}}$) of in equation \eqref{eq:recon_loss} are
assigned with $1$ and $2$ separately. It takes around 18 hours to train the whole
model on two NVIDIA A100 GPUs with 40GB memory.

\begin{table*}[t]
    \centering
    \setlength{\tabcolsep}{4pt}
    \caption{Comparison of gaze and head redirection on ETH-XGaze, ColumbiaGaze, and MPIIFaceGaze datasets. Lower values are better for Gaze, Head, and LPIPS, while higher values are better for ID.
    Our model demonstrates a consistent superiority over other SOTA methods across all key metrics on the ColumbiaGaze and MPIIFaceGaze datasets. These results highlight the robustness and adaptability of our model in effectively executing gaze redirection tasks, even when applied to diverse datasets.}
    \begin{tabular}{l|cccc||cccc|cccc}
        \toprule
        & \multicolumn{4}{c}{\textbf{ETH-XGaze}} & \multicolumn{4}{c}{\textbf{ColumbiaGaze}} & \multicolumn{4}{c}{\textbf{MPIIFaceGaze}} \\
        & \textbf{LPIPS}$\downarrow$ & \textbf{ID}$\uparrow$ & \textbf{Gaze}$\downarrow$ & \textbf{Head}$\downarrow$ 
        & \textbf{LPIPS}$\downarrow$ & \textbf{ID}$\uparrow$ & \textbf{Gaze}$\downarrow$ & \textbf{Head}$\downarrow$ 
        & \textbf{LPIPS}$\downarrow$ & \textbf{ID}$\uparrow$ & \textbf{Gaze}$\downarrow$ & \textbf{Head}$\downarrow$ \\   \midrule
        ST-ED & 0.300 & 24.347 & 16.217 & 13.153 
             & 0.413 & 6.384 & 17.887 & 14.693 
             & 0.288 & 10.677 & 14.796 & 11.893 \\
        HeadNeRF & 0.294 & 46.126 & 12.117 & 4.275 
                 & 0.349 & 23.579 & 15.250 & 6.255 
                 & 0.288 & 31.877 & 14.320 & 9.372 \\
        GazeNeRF & 0.291 & 45.207 & \textbf{6.944} & \textbf{3.470} 
                 & 0.352 & 23.157 & 9.464 & 3.811 
                 & 0.272 & 30.981 & 14.933 & 7.118 \\
        \rowcolor[rgb]{0.902,0.902,0.902} RTGaze & \textbf{0.262} & \textbf{60.708} & 9.047 & 3.631 
                 & \textbf{0.249} & \textbf{61.765} & \textbf{7.625} & \textbf{3.326} 
                 & \textbf{0.251} & \textbf{46.098} & \textbf{9.409} & \textbf{6.444} \\
        \bottomrule
    \end{tabular}
    
    \label{tab:accuracy}
\end{table*}

\begin{figure*}[t]
    \centering
    \includegraphics[width=1\linewidth]{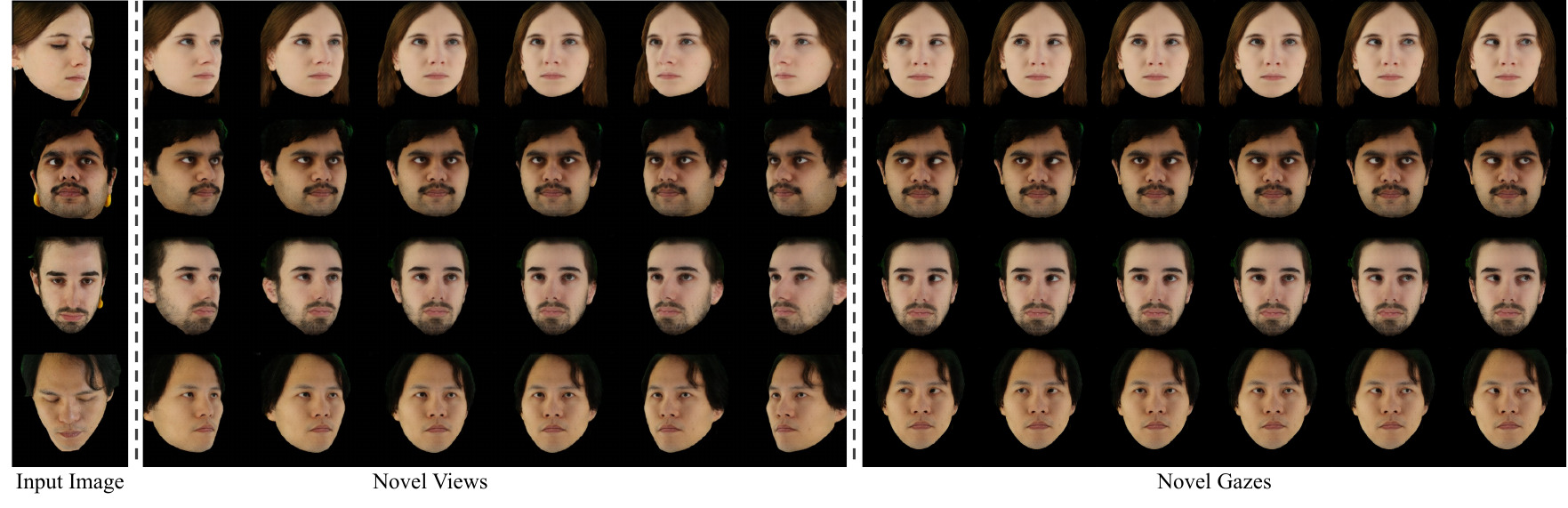}
    \caption{Visualization of generated results under novel views and gazes. Our
    model is able to generate 3D faces with controllable gazes using one single image
    as input. It can generate photorealistic face images in a large range of
    head pose and gaze directions. The results under novel views show that our model
    keeps good 3D consistency in the generation process. Its ability to generate
    consistent gaze images is also demonstrated by the results under novel gazes.
    Please zoom in for better visualization.
    }
    \label{fig:gaze-head}
\end{figure*}

\noindent \textbf{Evaluation Metrics.} We evaluate our model with various metrics regarding model efficiency, generated image quality, redirection accuracy, and identity preservation. 
To evaluate the efficiency of models, we report inference time (including encoding time and rendering time) measured on a single NVIDIA 3090 GPU in the inference stage with an average of 100 samples. 
To evaluate the quality of generated image, we report four widely used metrics including
Structure Similarity Index (SSIM) \cite{wang2004image}, Peak Signal-to-Noise
Ratio (PSNR), Learned Perceptual Image Patch Similarity (LPIPS) \cite{zhang2018unreasonable},
and Fr\'echet Inception Distance (FID).
To evaluate the accuracy of gaze redirection, we report gaze error and head error in degrees.
Identity similarity (ID) is assessed using the face recognition model from FaceX-Zoo \cite{wang2021facex}. This model evaluates the discrepancies in identity between the redirected images and the corresponding ground truth images.

\noindent \textbf{Baseline Methods.} We compare our model against the state-of-the-art gaze
redirection methods including 2D-based method ST-ED \cite{zheng2020self} and 3D-based
method GazeNeRF \cite{ruzzi2022gazenerf}. ST-ED realizes gaze redirection on
full-face images by disentangling latent vectors with a novel self-transforming
encoder-decoder architecture. GazeNeRF disentangles eye and face with two-stream
MLPs and achieves 3D-aware gaze redirection based on NeRF representation. 
We also compare our model with HeadNeRF \cite{hong2022headnerf}, a state-of-the-art NeRF-based 3D portrait generation model. 
It is adapted to gaze redirection task by simply adding two-dimension gaze labels as additional input.

\subsection{Quantitative Comparison in Image Generation}
We show the quantitative results of the comparison with
SOTA methods in Table \ref{tab:sota}.
We evaluate our model against other state-of-the-art
methods in terms of generated image quality, using widely used metrics including
SSIM, PSNR, LPIPS, and FID. To ensure fairness, we only compare our model with other
3D-based methods on inference speed, examining encoding time, rendering time, and total inference time. Image
quality is assessed on the personalized test set from the ETH-XGaze dataset, and
inference speed is measured by averaging results from $100$ samples on a single
NVIDIA 3090 GPU. 

For inference speed, RTGaze achieves real-time performance in
both encoding and rendering stages, with a total processing time of $61ms$ per image.
This is attributed to the efficient triplane-based lightweight module distilled
from a pre-trained 3D GAN \cite{trevithick2023real}, as well as the avoidance of
the inversion process by requiring only a single image as input. In contrast, both
HeadNeRF and GazeNeRF are based on the same parametric head model with NeRF representation.
Their inputs are parameters of a specific head instead of images. Therefore,
they have to conduct an inversion process to update the parameters with the
input image, which takes a great amount of time like one minute. They suffer
from slower encoding times due to the involved inversion process. Regarding image
quality, RTGaze beats the other SOTA methods on most metrics (PSNR, LPIPS, FID)
and achieves a comparable result on SSIM. Notably, our model outperforms other
methods on FID by a large margin.

\begin{table}[t]
    \centering
    \setlength{\tabcolsep}{2.5pt} 
    \caption{Ablation study on gaze prompt injection. The results indicate that injecting gaze prompt into low frequency feature cannot achieve competitive performance. }
    \begin{tabular}{ccccc}
        \hline
        Injecting into & \textbf{FID} $\downarrow$ & \textbf{ID} $\uparrow$ & \textbf{Gaze} $\downarrow$ & \textbf{Head} $\downarrow$ \\
        \hline
         Low-Frequency Feature  & $67.298$ & $38.517$ & $18.973$ & $5.409$ \\
        \rowcolor[rgb]{0.902,0.902,0.902} High-Frequency Feature & $\textbf{38.346}$ & $\textbf{60.708}$ & $\textbf{9.047}$ & $\textbf{3.631}$ \\
        \hline
    \end{tabular}
    \label{tab:ablationfeature}
\end{table}

\begin{table}[t]
    \centering
    \setlength{\tabcolsep}{5pt}
    \caption{Ablation study on loss functions. Note that, $\mathcal{L}_{\mathcal{D}}$ denotes the inclusion of 3D face prior distillation. }
    \begin{tabular}{lcccc}
        \hline
        & \textbf{FID} $\downarrow$ & \textbf{ID} $\uparrow$ & \textbf{Gaze} $\downarrow$ & \textbf{Head} $\downarrow$ \\
        \hline
        $\mathcal{L}_{\mathcal{R}}$ & $101.053$ & $47.251$ & $9.332$ & $4.208$ \\
        $\mathcal{L}_{\mathcal{R}}$ + $\mathcal{L}_{\mathcal{P}}$ & $54.682$ & $52.518$ & $10.911$ & $3.700$ \\
        \rowcolor[rgb]{0.902,0.902,0.902} $\mathcal{L}_{\mathcal{R}}$ + $\mathcal{L}_{\mathcal{P}}$ + $\mathcal{L}_{\mathcal{D}}$ & $\textbf{38.346}$ & $\textbf{60.708}$ & $\textbf{9.047}$ & $\textbf{3.631}$ \\
        \hline
    \end{tabular}
    \label{tab:ablation}
\end{table}

\subsection{Qualitative Comparison in Image Generation}

We show the qualitative results of the comparison with
SOTA methods in Fig. \ref{fig:sota}. Following GazeNeRF \cite{ruzzi2022gazenerf},
we pair the images with the different gazes from the personalized test set of
ETH-XGaze to get the input and target images. 
Our model takes an image and a target gaze label as inputs, generating a photorealistic gaze-adjusted image.

As shown in Fig. \ref{fig:sota}, ST-ED \cite{zheng2020self} suffers from
preserving identity information tending to generate similar faces with different
inputs. Besides, the results from ST-ED preserve the unmasked green background
by mistake which is barely found in 3D-based methods. 2D-based methods only learn a mapping from the input image and gaze label
to the target image, while 3D-based methods are trained to build 3D face
representations by integrating extensive multi-view information. It demonstrates the
robustness of 3D-based methods in handling defective inputs.  GazeNeRF \cite{ruzzi2022gazenerf} generates gaze-redirected
images whose gazes are aligned with target images, while it struggles to
preserve more facial details including facial texture and fine-grained hair. In contrast, our model can generate
photorealistic face images with extensive details while maintaining the ability
to redirect gaze accurately. Notably, our model works in real time which
outperforms all the existing 3D-aware methods.

\subsection{Gaze Redirection Accuracy Evaluation}
We further assess the performance of our model on the ETH-XGaze, ColumbiaGaze and MPIIFaceGaze datasets regarding redirection accuracy and identity preservation. We compare our model with ST-ED \cite{zheng2020self}, HeadNeRF \cite{hong2022headnerf}, and GazeNeRF \cite{ruzzi2022gazenerf} in terms of gaze error, head error, LPIPS, and identity similarity.
The detailed results of this evaluation are presented in Table \ref{tab:accuracy}. 
Our model outperforms the competing methods regarding image quality (LPIPS) and identity preservation (ID) and achieves comparable redirection accuracy on ETH-XGaze dataset.
Notably, our model consistently outperforms competing methods across all key metrics on ColumbiaGaze and MPIIFaceGaze datasets. These findings underscore the robustness and adaptability of our model in performing gaze redirection tasks, even when applied to diverse datasets.

\subsection{Face Rendering under Novel Views and Gazes}

To showcase the effectiveness of our model in generating 3D-consistent results
and achieving consistent gaze redirection, we show the visualization of face rendering
under novel views and gazes in Fig. \ref{fig:gaze-head}. We set the gaze as looking
forward during the generation under novel views and interpolate the gaze from left
to right under a frontal view in the generation under novel gazes. The results
demonstrate that our model can generate face images with strong 3D consistency
and enables smooth and coherent gaze interpolation. The good performance relies on
our expressive gaze-controllable facial representation and the simple but effective
face geometric prior distillation. It is important to note that our model allows the generation of photorealistic face images across a large range of head poses and gaze directions. 

\subsection{Ablation Study}

\begin{figure}
    \centering
    \includegraphics[width=1.01\linewidth]{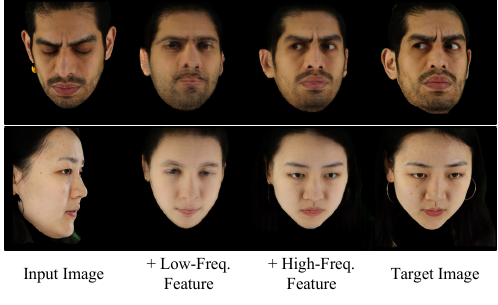}
    \caption{Visualization of ablation on fusion feature choices. 
    The findings suggest that solely relying on low-frequency geometric features leads to blurriness and inaccurate gaze redirection. Conversely, combining high-frequency appearance features with gaze embedding maintains facial structure while enabling efficient gaze redirection.}
    \label{fig:ablation_feature}
\end{figure}

\begin{figure}[t]
    \centering
    \includegraphics[width=1\linewidth]{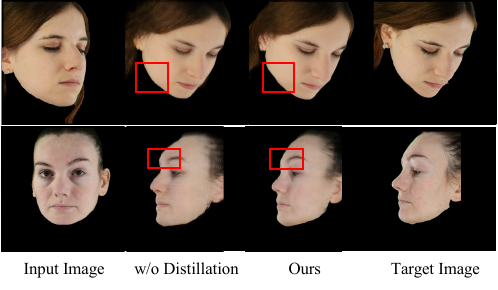}
    \caption{Visualization of ablation on 3D face prior distillation. The results without the distillation show shape distortions, whereas the model utilizing the 3D prior accurately reconstructs the 3D shape.}
    \label{fig:ablation_depth}
\end{figure}

\textbf{Gaze-Controllable Facial Representation.} RTGaze extracts high-frequency and low-frequency features, and the gaze prompt injection module inject gaze prompt embedding with high-frequency feature. 
We first perform ablation study on the choices of
features for gaze prompt injection.
The result is shown in Table \ref{tab:ablationfeature}, where injection with low-frequency feature cannot achieve competitive performance. 
We also try to inject gaze prompt into both low-frequency and high-frequency features, but the training fails to converge due to the difficulty of modifying appearance and geometry simultaneously.
We present the generated images with the fusion of low-frequency features in Fig. \ref{fig:ablation_feature}. The results exhibit overall blurriness, and the gaze redirection fails to produce accurate outcomes. When the gaze embedding is fused with low-frequency features, the geometry of the 3D face entangled  with the input gaze labels. Ideally, the model should modify only the geometry
around the eye region; however, without specific geometric constraints, the model
struggles to focus on the eye region alone. Instead, it tends to alter the entire
face, leading to noticeable instability in the generated results. In our approach,
we fuse the high-frequency features with the gaze embedding while keeping the
geometric features unchanged. This enables the model to perform gaze redirection
by adjusting only the appearance of the eye region, ensuring a stable face shape
throughout the process.

\noindent \textbf{3D Face Prior Distillation.} 
We also conduct an ablation study on the 3D face prior distillation and the proposed loss functions.
Results are shown in Table \ref{tab:ablation}. 
$\mathcal{L}_{\mathcal{D}}$ denotes the inclusion of 3D face prior distillation in our method.
The results demonstrate that incorporating 3D face prior distillation and proposed loss functions significantly improves the performance of the model.
We also show the qualitative results with and without 3D face prior distillation in Fig. \ref{fig:ablation_depth}. The results without 3D prior exhibit distortions in shape, while the model with 3D prior successfully reconstructs the 3D shape of the input face. This demonstrates that the chosen 3D GAN prior provides effective information on 3D face shape, improving the final generation performance.

\section{Conclusion}

We propose RTGaze, a real-time 3D-aware gaze redirection method from a single image. 
Our model achieves real-time inference by employing an expressive gaze-controllable facial representation to directly fuse gaze prompt into image space and distilling the prior knowledge from 3D GANs into a lightweight module. 
Benefiting from the gaze-controllable facial representation and the face geometric prior distillation, our model realizes accurate gaze redirection while maintaining superior image quality. 
With its exceptional real-time performance and high-quality generation, our model holds great potential for numerous downstream applications, particularly in scenarios with high real-time requirements.

\section{Acknowledgments}
This work was supported by Institute of Information \& communications Technology Planning \& Evaluation (IITP) grant funded by the Korea government(MSIT) (No.2022-0-00608, Artificial intelligence research about multi-modal interactions for empathetic conversations with humans).
This research was also supported by National Natural Science Foundation of China (Grant No. 62302252).
The research utilized the Baskerville Tier 2 HPC service (https://www.baskerville.ac.uk/) funded by the Engineering and Physical Sciences Research Council (EPSRC) and UKRI through the World Class Labs scheme (EP/T022221/1) and the Digital Research Infrastructure programme (EP/W032244/1) operated by Advanced Research Computing at the University of Birmingham. 
Hengfei Wang was supported by China Scholarship Council Grant No.202006210057. 

\bibliography{main}

\end{document}